\documentclass[letterpaper, 10 pt, conference]{ieeeconf}  

\IEEEoverridecommandlockouts                              
                                                        
\usepackage{graphicx} 
\usepackage{epsfig} 
\usepackage{mathptmx} 
\usepackage{times} 
\usepackage{amsmath} 
\usepackage{amssymb}  

\usepackage{xcolor}
\usepackage{subfigure}

\usepackage{algorithm}
\usepackage{algorithmic}
\usepackage{todonotes}

\usepackage{hyperref}

\newcommand{\comPos}{\mathbf{c}}
\newcommand{\comVel}{\dot{\comPos}}
\newcommand{\comAcc}{\ddot{\comPos}}
\newcommand{\force}{\mathbf{f}}
\newcommand{\gravity}{\mathbf{g}}

\newcommand{\jointState}{\mathbf{q}}

\newcommand{\jointAcc}{\ddot{\jointState}}

\title{\LARGE \bf
Control of Humanoid in Multiple Fixed and Moving Unilateral Contacts
}

\author{Julien Roux$^{1}$, Saeid Samadi$^{1}$, Eisoku Kuroiwa$^{2}$, Takahide Yoshiike$^{2}$, and Abderrahmane Kheddar$^{3,1}$
\thanks{$^{1}$CNRS-University of Montpellier, LIRMM, Montpellier, France.
	Corresponding author: {\tt\footnotesize julien.roux@lirmm.fr}}
\thanks{$^{2}$Honda Research Institute, Tokyo, Japan.}%
\thanks{$^{3}$CNRS-AIST Joint Robotics Laboratory, IRL3218, Tsukuba, Japan.}
}
\begin{document}

\maketitle
\thispagestyle{empty}
\pagestyle{empty}

\begin{abstract}
Enforcing balance of multi-limbed robots in multiple non-coplanar unilateral contact settings is challenging when a subset of such contacts are also induced in motion tasks.
The first contribution of this paper is in enhancing the computational performance of state-of-the-art geometric center-of-mass inclusion-based balance method to be integrated online as part of a task-space whole-body control framework. 
As a consequence, our second contribution lies in integrating such a balance region with relevant contact force distribution without pre-computing a target center-of-mass. This last feature is essential to leave the latter with freedom to better comply with other existing tasks that are not captured in classical two-level approaches. We assess the performance of our proposed method through experiments using the HRP-4 humanoid robot.
\end{abstract}

\section{Introduction}

A critical issue in planning~\cite{bouyarmane2009icra} and controlling~\cite{bouyarmane2018tac} humanoid robots in multi-contact settings~\cite{samadi2021ral} is to devise dynamic balance conditions accounting for various types of unilateral contacts~\cite{bouyarmane2018springer}. Both~\cite{bretl2008tro} and~\cite{audren2018tro} considered multi-contact equilibrium criterion featuring non-coplanar contacts with friction in static and dynamic configurations, respectively. A balance region is defined such that the Center of Mass (CoM) of the robot should stay inside during its motion to avoid falling. The balance region computation is based on centroidal dynamics derived from the Newton-Euler equations with the friction cones inequalities constraints. They both used an iterative ray shooting (projection) algorithm to compute an inclusion region for the Center of Mass (CoM) position. The CoM acceleration is considered null in~\cite{bretl2008tro}, and bounded in a convex-set in~\cite{audren2018tro}. We recall these methods in Section~\ref{sec:background}.

Other methods to construct similar regions are based on sampling~\cite{sentis2010tro}, approximation~\cite{nozawa2016humanoids} or learning~\cite{carpentier2017rss}.
However, such computed ``regions'' are used for planning CoM trajectories~\cite{audren2017humanoids,orsolino2020tro,focchi2017autonomousrobots} or for testing the feasibility of CoM positions~\cite{delprete2016icra} but not directly online.
Indeed, the main disadvantage of geometric balance regions lie in their computational cost; they are not useful when any contact location undergo motion tasks, e.g. pushing, sliding, etc. The main contribution of our paper is to propose a faster version of the iterative algorithms from~\cite{audren2018tro} to use it as a balance criterion when contacts move, see Section~\ref{sec:improvements}.

\begin{figure}[!tb]
  \centering
  \includegraphics[width=0.49\columnwidth]{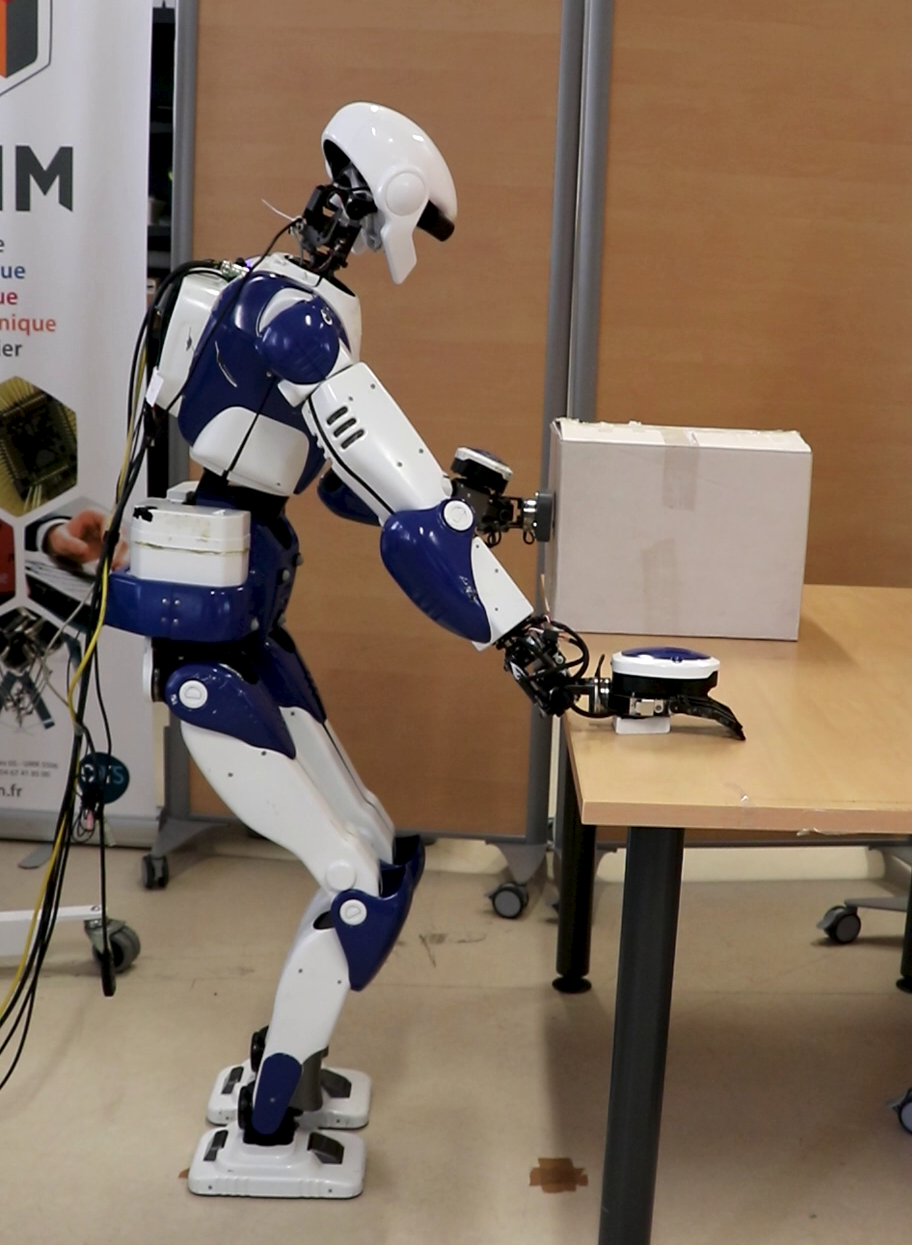}
  \includegraphics[width=0.49\columnwidth]{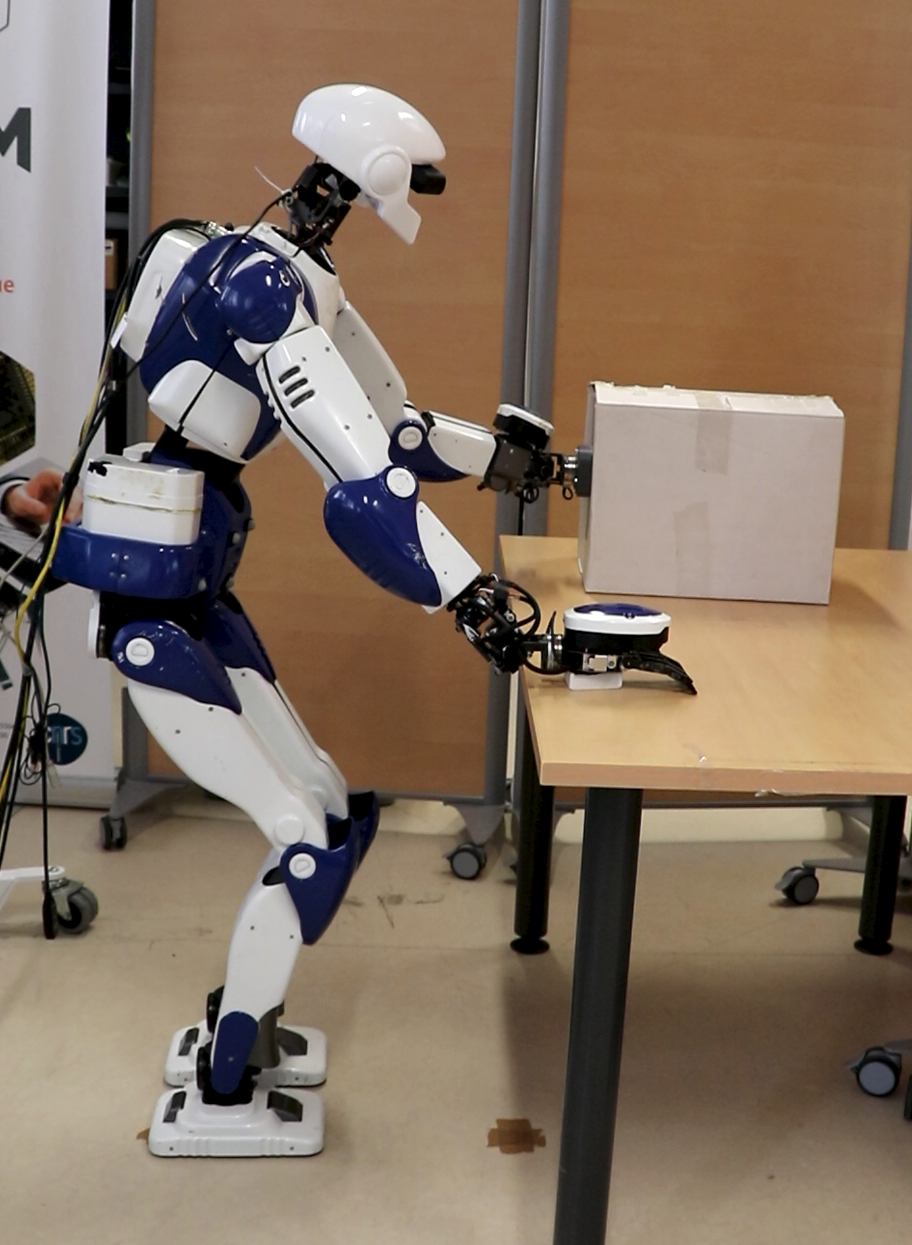}
  \caption{HRP-4 pushing a $12.5$~kg box. It produced a force of up to $30$~N horizontally with its left hand while controlling the remaining fixed contacts during the pushing phase.}
  \label{fig:humanRobotInteraction}
\end{figure}

In our recent work~\cite{samadi2021ral}, we proposed an approximation of the exact geometric (polyhedron) balance region using the Chebychev center (enclosed hyper-sphere), which allows fast computations. This method is successfully applied to multiple fixed-and-sliding contacts, yet it has two shortcomings: (i) the computed region is conservative (not critical in some use-cases), (ii) it requires two-level computations: the first one (a quadratic program) uses centroidal dynamics to compute the contact force distributions (accounting for any desired force at a given contact) and the CoM placement under balance constraints; the second level (another quadratic program) is the whole-body controller~\cite{bouyarmane2019tro}. The latter achieves computed CoM and forces as additional desired tasks. These two levels are operating in closed-loop. However, the centroidal dynamic computation is blind to other tasks that are to be achieved by the whole-body controller. Subsequently, the tracking of the computed balance CoM interferes with any other (postural) tasks that induce changes in the CoM, e.g. reaching the target with an end-effector, avoiding collisions, gazing, etc. The generated humanoid robot motion is somewhat awkward in many situations. In this paper, both drawbacks are avoided since the balance shape is computed exactly and is passed to the whole-body controller to constrain to CoM within the computed region, without explicitly specifying its target. Experiments demonstrating the capabilities of our approach are shown in Section~\ref{sec:experiments}; see Fig.~\ref{fig:humanRobotInteraction} illustrating the experimental set-up. 

\section{Centroidal Model and CoM Balance Region}
\label{sec:background}

\subsection{Centroidal Model}

Let $\comPos\in\mathbb{R}^3$ be the CoM position of a floating base multi-limb robot and $m>0$ its mass. The robot interacts with its environment through a contact set $\mathcal{F}$ of $n$ unilateral contact points. For each contact point $i$, $\mathbf{r}_i\in\mathbb{R}^3$ is its contact position, $\mathbf{u}_i\in\mathbb{R}^3$ its normal, $\force_i\in\mathbb{R}^3$ its force applied on the robot and $\mu_i$ its friction coefficient. Newton-Euler equations governing the CoM motion writes,
\begin{equation}
  \label{eq:sum_forces}
  \sum_{i=1}^{n} \force_i + m\gravity = m\comAcc
\end{equation}
\begin{equation}
  \label{eq:sum_moment}
    \sum_{i=1}^{n} \mathbf{r}_i \times \force_i +  \comPos\times m \gravity = \comPos\times m\comAcc
\end{equation}
where $\gravity$ is the gravity vector; the moment equation is expressed w.r.t. the world's origin and assumes neglectible angular momentum.

The friction force for the contact point $i$ is given by:
\begin{equation}
  \label{eq:friction1}
  \mathbf{u}_i^T \force_i \geq 0
\end{equation}
\begin{equation}
  \label{eq:friction2}
  \|(\mathbb{I}-\mathbf{u}_i\mathbf{u}_i^T)\force_i \| \leq \mu_i \mathbf{u}_i^T \force_i
\end{equation}

An equilibrium can be described by the vector $(\force^T, \comPos^T)^T$ with $\force= (\force_1^T, \dots, \force_n^T )^T\in\mathbb{R}^{3n}$ such that equations~\eqref{eq:sum_forces} to~\eqref{eq:friction2} hold. Let $\mathcal{X}$ be the set of all equilibrium:
\begin{equation}
  \mathcal{X} = \left\{
  \begin{pmatrix}
    \force\\
    \comPos
  \end{pmatrix} \in\mathbb{R}^{3n}\times\mathbb{R}^3 \,\,\, |\,\,\, \eqref{eq:sum_forces} - \eqref{eq:friction2}
  \right\}
\end{equation}

Implementing $\mathcal{X}$ as constraints in planning or control frameworks enforces no sliding (even in a moving contact) and no contact break. Notice that the set of constraints is non-linear because of the cross product in~\eqref{eq:sum_moment}. We aim to compute fast the boundaries of the equilibrium set in the 3D CoM space, i.e., finding the edges of the convex set:
\begin{equation}
  \mathcal{Y} = \{\comPos\in\mathbb{R}^3 \,\,\, |\,\,\, \exists \, \force\in\mathbb{R}^{3n}, \,(\comPos, \force)\in \mathcal{X}\}\nonumber
\end{equation}


\begin{algorithm}[t]
  \caption{Iterative Projection Algorithm}
  \label{projection_pseudo_code}
  \begin{algorithmic}
    \REQUIRE Robot's contact set $\mathcal{F}$
    \REQUIRE $\cal V(x)$ computes volume of $\cal x$ 
    \STATE Initialize $\mathcal{Y}_{\text{inner}}$ and $\mathcal{Y}_{\text{outer}}$
    \WHILE {${\cal V}(\mathcal{Y}_{\text{outer}})-{\cal V}(\mathcal{Y}_{\text{inner}}) >\epsilon$}
    \STATE -- $\forall \text{ face } H_i \in \mathcal{Y}_{\text{inner}}$, compute ${\cal V}_i(\mathcal{Y}_{\text{outer}}) \cap H_i$
    \STATE -- face $i \leftarrow \max {\cal V}_i$.
    \STATE -- new search direction $\mathbf{d}^*$ normal to the face $i$.
    \STATE $*$ find CoM $\comPos^*$ from SOCP problem~\eqref{eq:SOCP} for $\mathbf{d}^*$:
    \begin{equation}
      \max_{\force, \comPos}\; \mathbf{d}^{*T} \comPos \quad \text{s.t.} \; 
      \begin{pmatrix}
        \force\\
        \comPos
      \end{pmatrix} \in \mathcal{X}
      \nonumber
    \end{equation}
    \STATE $+$ Update $\mathcal{Y}_{\text{inner}}$ by adding $\comPos^*$
    \STATE $\times$ Update $\mathcal{Y}_{\text{outer}}$ by adding the plane of normal $\mathbf{d}^*$ and  passing through $\comPos^*$
    \ENDWHILE
    \RETURN $\mathcal{Y}_{\text{inner}}$
  \end{algorithmic}
\end{algorithm}

\subsection{CoM balance region in multi-contact}
This problem is studied in~\cite{bretl2008tro} for static balance, i.e., $\comAcc=\mathbf{0}$. The resulting region does not depend on the CoM's height $c_z$, nor on the mass of the robot~\cite{caron2016humanoids}. The equilibrium region is infinite in the direction parallel to gravity (noted $z$) with a constant projection on the horizontal plane (this is because bounds on torques and kinematic constraints are not accounted for in the centroidal model). They proposed to compute the cross-section only. The resulting equilibrium region limits the CoM position in the horizontal plane. 

In~\cite{samy2017humanoids}, the static balance region is extended to account for the effector torque limits that is also implemented in~\cite{orsolino2020tro}. Using the projection algorithm in~\cite{bretl2008tro}, the balance region is computed at each control-loop iteration of their quadruped robot having at most four-point contacts. However, in almost all practical situations, the contacting surfaces cannot be approximated as a single point contact. Instead, the contacting surface can be modeled as a set of contact points representing the vertices of the contacting surface discrete boundaries~\cite{caron2015leveraging}.

In~\cite{audren2018tro}, the followings are considered: (i) the CoM acceleration is bounded by a predefined convex set; (ii) each component of the contact force is constrained to be lower than the weight of the robot ($\|\force_i\|_\infty\leq m\gravity$); and (iii) the CoM position is constrained in a sphere of radius $c_{\max}$. That case is referred to as the robust equilibrium case as the CoM acceleration can correspond to external perturbations. The resulting 3D polytope depends on the mass of the robot and is used to constrain the CoM of a humanoid robot in motion.

In these works, the equilibrium balance region is obtained through an iterative algorithm based on a cutting plane method to approach $\mathcal{Y}$ using two convex polytopes $\mathcal{Y}_{\text{inner}}$ and $\mathcal{Y}_{\text{outer}}$, such that $\mathcal{Y}_{\text{inner}}\subseteq\mathcal{Y}\subseteq\mathcal{Y}_{\text{outer}}$ and all vertices of $\mathcal{Y}_{\text{inner}}$ lie on the boundary of $\mathcal{Y}$ and all faces of $\mathcal{Y}_{\text{outer}}$ are supporting planes of $\mathcal{Y}$ at the vertices of $\mathcal{Y}_{\text{inner}}$.

\begin{figure}[!thb]
  \centering
  \includegraphics[width=7cm]{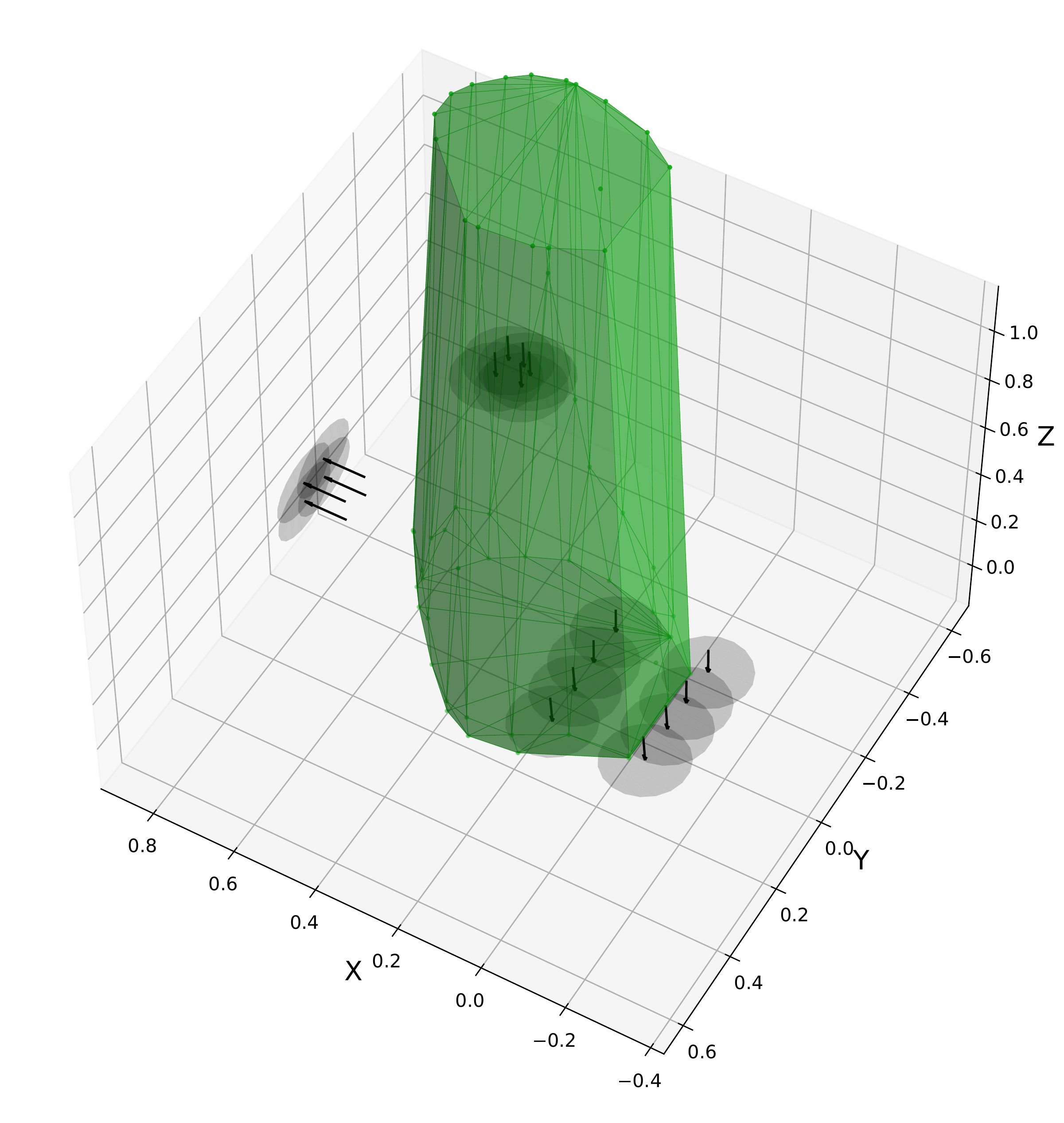}
  \caption{
  The balance CoM region for a robot in a contact configuration similar to Fig~\ref{fig:humanRobotInteraction}. The CoM acceleration is bounded to $\pm 0.4$~m.s$^{-2}$ in $x$-axis, $\pm 0.3$~m.s$^{-2}$ in $y$-axis and $-9.81 \pm 0.3$~m.s$^{-2}$ in $z$-axis. The black arrows are the contact used for the computation, each one of them is a vertex of a contact surface. The region is truncated at $0 \leq z \leq 2$~m}
  \label{fig:equilibrium_regions}
\end{figure}

In order to find points that are on the boundary of $\mathcal{Y}$ and the corresponding supporting planes, the following Second Order Cone Programming (SOCP) problem is formulated:
\begin{eqnarray}
  \label{eq:SOCP}
  \max_{\comPos, \force}& \mathbf{d}^T \comPos \\
  \text{s.t.}& 
  \begin{pmatrix}
    \force \\
    \comPos
  \end{pmatrix}
  \in \mathcal{X}\nonumber
\end{eqnarray}
where $\mathbf{d}\in \mathbb{R}^3$ is a search direction. The iterative projection algorithm (Algorithm~\ref{projection_pseudo_code}) returns the inner approximation polytope $\mathcal{Y}_{\text{inner}}$ of the convex set $\mathcal{Y}$, see e.g. Fig.~\ref{fig:equilibrium_regions}. Then its H-Rep is employed to constrain the CoM position in a planning or control framework.

\section{Improving the Projection Algorithm}
\label{sec:improvements}
We describe the improvements made in the computation of the dynamic equilibrium region to achieve higher time performances. The proposed modifications for Algorithm~\ref{projection_pseudo_code} are the followings:
\begin{itemize}
\item Step $*$: Linearizing the constraints and use Linear Programming (LP) instead of SOCP (Section~\ref{sec:LPvsSOCP});
\item Computing support functions instead of volumes $\cal V$ (Section~\ref{sec:SupportFunctions});
\item Step $+$: Warm starting the inner update (Section~\ref{sec:update_inner_outer});
\item Step $\times$: Warm starting the outer update (Section~\ref{sec:update_inner_outer}).
\end{itemize}

\subsection{Linearization of the constraints}
\label{sec:LPvsSOCP}
Linearization is a rather common practice and has already been applied to Algorithm~\ref{projection_pseudo_code} in~\cite{caron2016humanoids,orsolino2020tro}. Linearizing the constraint~\eqref{eq:friction2} allows using Linear Programming (LP) instead of Second Order Cone Programming (SOCP) proposed in~\cite{bretl2008tro,audren2018tro}. The linearized friction cones are given by:
\begin{equation}
  \label{eq:FrictionLinear}
  \frac{(\mathbb{I}-\mathbf{u}_i\mathbf{u}_i^T)\force_i}{\mu_i \mathbf{u}_i^T \force_i} \in \mathcal{P}_n
\end{equation}
where $\mathcal{P}_n$ is the regular $n$-sided polygon inscribed in the unit circle. 
In addition, the maximal normal contact forces are bounded by $f_{\text{max}}$:
\begin{equation}
  \label{eq:ForcesUpperBound}
  \mathbf{u}_i^T \mathbf{f_i} \leq f_{\text{max}}
\end{equation} $\mathbf{u}_i$ is the normal of the $i$-th contact. $f_{\max}$ depends on the configuration as it is related to joint torques limits~\cite{orsolino2020tro}. In this study, $f_{\max}$ is taken constant.

Moreover, contrarily to bounding the CoM to a sphere~\cite{audren2018tro}, it can be determined using linear constraints:
\begin{equation}
  A\comPos\leq \mathbf{b}
\end{equation}

This new constraint results in a convex set $\mathcal{X}_{\text{linear}} \subseteq \mathcal{X}$:
\begin{equation}
  \mathcal{X}_{\text{linear}} = \left\{
    \begin{pmatrix}
      \force\\
      \comPos
    \end{pmatrix} \in \mathbb{R}^{3n+3} \left| \;\eqref{eq:sum_forces},\eqref{eq:sum_moment},\eqref{eq:friction1},\eqref{eq:FrictionLinear} \;\right.\right\}
\end{equation}
The projection in the CoM position space becomes:
\begin{equation}
  \mathcal{Y}_{\text{linear}} = \left\{\comPos\in\mathbb{R}^3\,\,\,| \,\,\,\exists \force\in\mathbb{R}^{3n},\,
  \begin{pmatrix}
    \force\\
    \comPos
  \end{pmatrix}\in\mathcal{X}_{\text{linear}}
  \right\}
\end{equation}
Constraint~\eqref{eq:FrictionLinear} being stricter than constraint~\eqref{eq:friction2} (i.e. for all $\force$ such that~\eqref{eq:FrictionLinear} holds then~\eqref{eq:friction2} holds as well), we have $\mathcal{X}_{\text{linear}}\subset\mathcal{X}$ and $\mathcal{Y}_{\text{linear}}\subset\mathcal{Y}$.
Finally we replace the SOCP with the following LP:
\begin{eqnarray}
  \label{eq:LP}
  \max_{\comPos, \force}& \mathbf{d}^T \comPos \\
  \text{s.t.}& \begin{pmatrix}
    \force \\
    \comPos
  \end{pmatrix} \in \mathcal{X}_{\text{linear}}\nonumber
\end{eqnarray}

\subsection{Support Point and Support Functions}
\label{sec:SupportFunctions}
The first step of each iteration in~\cite{audren2018tro} consists of computing the volumes cut off $\mathcal{Y}_{\rm outer}$ by the faces of $\mathcal{Y}_{\rm inner}$. However, computing the volume of a polytope is time-consuming~\cite{bueler2000springer}.

Let $H_i$ be a face of $\mathcal{Y}_{\text{inner}}$. $\mathbf{d}_{H_i}\in\mathbb{R}^3$ and $b_{H_i}\in\mathbb{R}$ are respectively the normal and offset of face $H_i$ such that
\begin{equation}
  \mathcal{Y}_{\text{inner}} \subset H^-_i = \{ \comPos\in\mathbb{R}^3 \,\,\, | \,\,\, \mathbf{d}_{H_i}^T  \comPos \leq b_{H_i}\}
\end{equation}

$\comPos_{H_i}\in\mathcal{Y}_{\rm outer}$ is a \textit{support point} w.r.t. face $H_i$ if
\begin{equation}
  \label{eq:support_point}
  \forall \comPos\in\mathcal{Y}_{\rm outer},\quad \mathbf{d}_{H_i}^T  \comPos \leq \mathbf{d}_{H_i}^T  \comPos_{H_i}
\end{equation}
$\comPos_{H_i}\in\mathcal{Y}_{\rm outer}$ is the furthest point outside $H^-_i$ and inside $\mathcal{Y}_{\rm outer}$.
Then, the \textit{support function} $s_i$ is the euclidean distance from a face of $\mathcal{Y}_{\rm inner}$ to its corresponding support point:
\begin{equation}
  s_i = \mathbf{d}_{H_i}^T \comPos_{H_i} - b_{H_i}
\end{equation}

As $\mathcal{Y}_{\rm outer}$ is a bounded convex polytope described by a finite amount of vertices, then $\comPos_{H_i}$ is one of its vertices. The search direction for the next iteration of the algorithm is the normal of the inner approximation's face with the maximum support function.

In~\cite{bretl2008tro,audren2018tro} proof of convergence using area/volume difference respectively between the inner and outer regions are provided. They use this difference as a stopping criterion. Instead, we propose a different metric that uses support functions and surfaces area instead of volumes:
\begin{equation}
  \sum_{i\in [1, m]}\frac{1}{3} s_i \mathcal{A}_i
\end{equation} where $m$ is the number of faces of $\mathcal{Y}_{\rm inner}$, $s_i$ and $\mathcal{A}_i$ are respectively the support function and the area of $i$-th face. For each face $i$,  $\frac{1}{3} s_i \mathcal{A}_i$ corresponds to the volume of the tetrahedron of the base and height of the support function.

This method is at a time fast and easy to compute. However, as only a lower bound of the error is provided, it cannot be used to prove convergence explicitly. Nonetheless, extensive simulations show good results and similar rates of convergence w.r.t the previous methods. 

\subsection{Updating the inner and outer approximations} \label{sec:update_inner_outer}
Updating the inner and outer polytopes corresponds respectively to an iterative face enumeration and iterative vertex enumeration problems. 
As the inner (resp. outer) polytope is convex, each face (resp. vertice) to update share at least one common edge with another face (resp. vertex) to update. Thus, in the general case, the complex step is to find the first face (resp. vertex) to update. In our implementation, the first face (resp. vertex) to update is given by the search face (resp. support point).

\subsection{Performance Comparison}
\label{sec:timings}

To compare fairly our method to the original one~\cite{audren2018tro}, both are implemented in Python. The contact sets obtained during the simulation of a scenario of the robot pushing a box is used. In total, $893$ different contact sets were used.
The number of contacts is from 2 to 4 surface contacts corresponding to 8 to 17 point contacts. For instance Fig~\ref{fig:humanRobotInteraction} represents a robot with 4 surface contact, but the model used for the computation consists of 17 contacts points as shown in Fig~\ref{fig:equilibrium_regions}. The computation performances depend on the number of contact points; see Table~\ref{tab:timmings}.
\begin{table}[tbhp]
  \centering
  \caption{Timings in ms}
  \label{tab:timmings}
  \begin{tabular}{|c||c|c|c||c|c|c|}
    \hline
    Method & \multicolumn{3}{c||}{SOCP and Volumes} & \multicolumn{3}{c|}{Linearized and Support}\\
    \hline \hline
    Nbr of contacts & 8 & 13 & 17 & 8 & 13 & 17 \\
    \hline
    Solve & 438 & 1219 & 1977 & 44 & 202 & 332\\ \hline
    Measure & 24 & 35 & 32 & 3 & 7 & 7 \\ \hline
    Update & 9 & 12 & 12 & 6 & 11 & 12 \\ \hline \hline
    Total & 473 & 1268 & 2022 & 54 & 222 & 352 \\ \hline
  \end{tabular}  
\end{table}

First of all, our implementation of the original method in~\cite{audren2018tro} (SOCP and Volumes in Table~\ref{tab:timmings}) show results that are even slightly better than those reported in~\cite{audren2018tro}, which is in favor of a fair comparison.

Overall, our proposed method is 6 to 9 times faster w.r.t the original one. The main gain comes from the use of LP instead of SOCP. Changing the measure decreased its computation time from 5 to 8 times. As for updating the inner and outer approximation, our method gives slightly better results w.r.t the original one. Finally, for the final version, a {\tt C++} implementation of the algorithm is used, and the computation time performances are much better: $25$~ms for two surface contacts, $80$~ms for three surface contacts, and $175$~ms for four surface contact.
\begin{figure}[tbhp]
  \centering
  \includegraphics[width=\columnwidth]{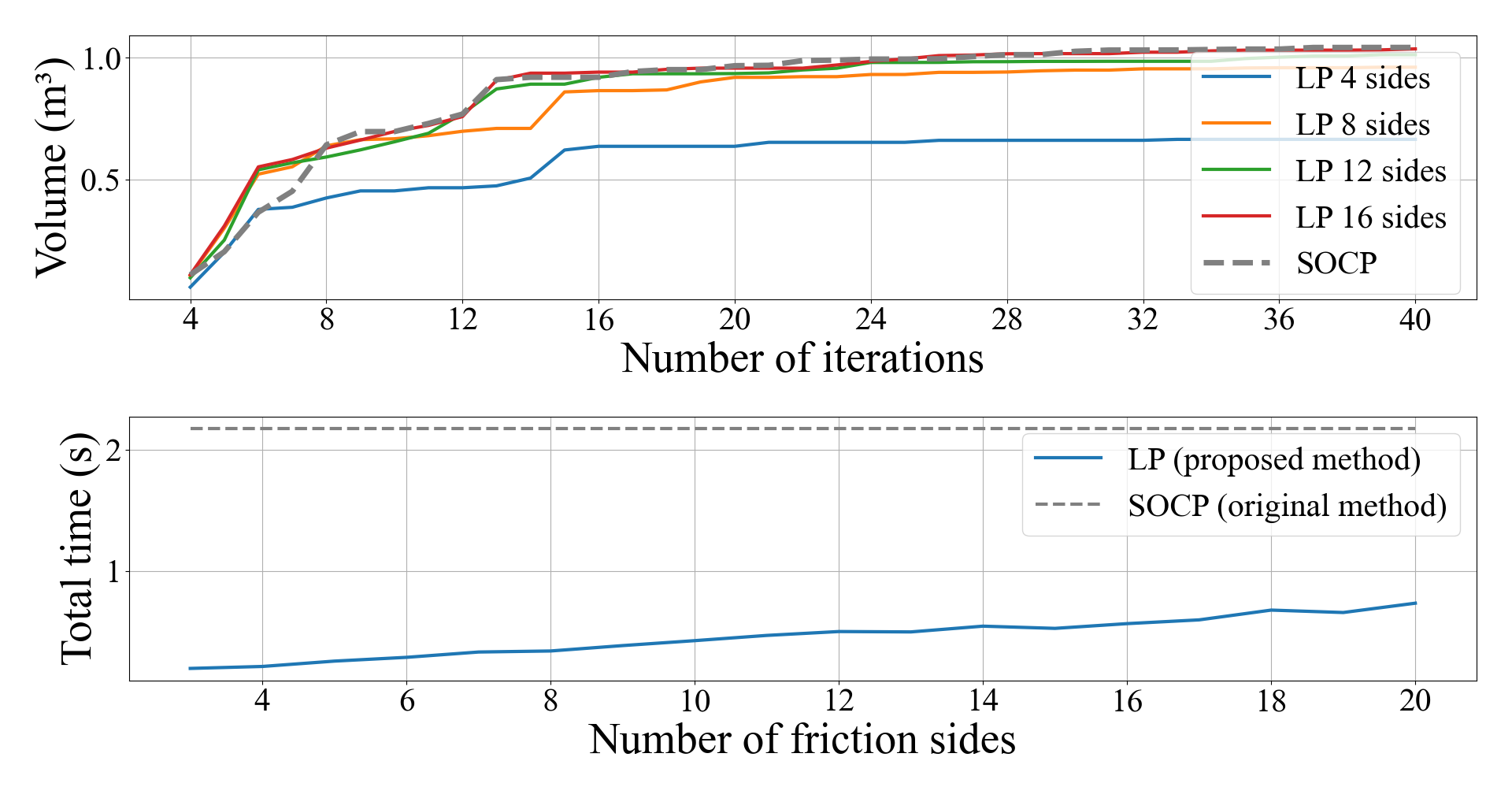}
  \caption{Impact of linearization; Top: evolution of the volume of the approximation w.r.t. the number of iteration of the projection algorithm for different number of sides for the linearization of the friction cone. Bottom: Evolution of the computation time w.r.t. the number of friction sides for the linearization of the friction cone}
  \label{fig:linearization_impact}
\end{figure}

Furthermore, The region generated with linearized friction cones tends to be smaller in volume compared to the original one when the number iterations increase (Fig.~\ref{fig:linearization_impact} top). Thus the result from our method is slightly more conservative than the previous methods.
Increasing the number of side for the linearization of the friction cones make the linearized region tend to the non-linearized one but the computation time will increase (Fig.~\ref{fig:linearization_impact} bottom). 
We found that a number of side between $6$ and $8$ is a good compromise.

The {\tt C++} library implementing the iterative projection algorithm is available online\footnotemark\footnotetext{\url{https://github.com/JulienRo7/stabiliplus}}

\section{Integration in the Whole-Body Control}
\label{sec:control_strategy}
Our proposed CoM balance region computation is introduced as a constraint in a task-space whole-body control framework. The latter was successfully used in highly complex scenarios~\cite{kheddar2019ram}. However, as discussed in Section~\ref{sec:control_strategy_context}, this constraint alone does not warrant that contact forces regulation tasks are achieved.
Thus, in Section~\ref{sec:control_strategy_CoM_Admittance}, a CoM admittance task is introduced such that the CoM moves to a position allowing for better tracking of the desired contact forces.

\begin{figure}[tbhp]
  \centering
  \includegraphics{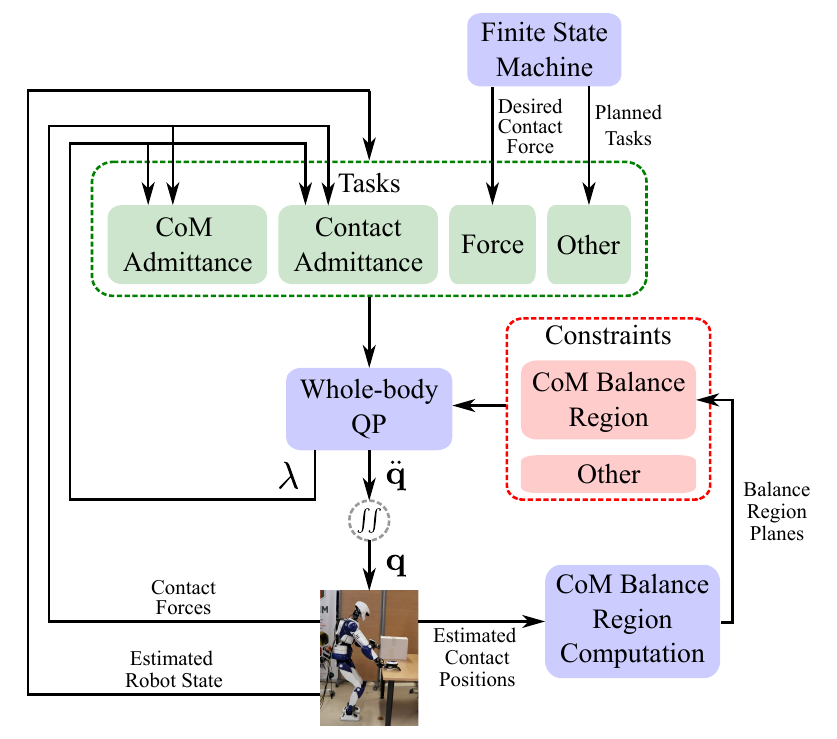}
  \caption{Tasks and constraints are formulated as function of the generalized acceleration coordinates $\jointAcc$ and contact force $\mathbf{\lambda}$. The targets of the tasks are given by a FSM or the WBQP's output from previous iteration and are compared to the estimated state of the robot. Using the estimated contact position the balance region is computed and used as constraints.}
  \label{fig:WBQP_graph}
\end{figure}

\subsection{Whole Body Balance Framework}
\label{sec:control_strategy_context}

Our task-space Whole-Body Quadratic Programming (WBQP) framework~\cite{vaillant2016springer} formulates quadratic task objectives and constraints as a function of the generalized acceleration coordinates and contact forces.
Figure~\ref{fig:WBQP_graph} illustrates our global framework: a Finite State Machine (FSM) schedules tasks on demand of the objectives and the used robot, and parametrize these tasks such as contact force, end effector position, posture... with their respective targets. As our robot is controlled in a high gain position, only the computed joint accelerations are used as commands after being integrated twice. Thus, for the robot to track the desired contact forces, we use admittance tasks as in~\cite{bouyarmane2019tro}.
The balance region is integrated as planar inequality constraints in the CoM position as in~\cite{Audren2016Humanoids}. Other concurrent constraints include joint limits (position and velocity), non-desired collision avoidance, and contact constraints.

The CoM is allowed to live freely within the computed balance region. In practice, a more refined control of CoM position is required, e.g.~\cite{samadi2021ral,abi-faraji2019ral,caron2019icra}.
Consider a task requiring the robot to apply a contact force $\force$ to its environment. To generate such a force and fulfill the dynamic constraint, the WBQP will adopt two strategies: (i) compensating $\force$ with internal forces and other contacts forces, and (ii) generating a CoM acceleration $\comAcc$ in the opposite direction of $\force$.
The former solution drives the contact wrenches close to their corresponding friction cones, leading to undesired contact slips. The latter can only be applied for a short amount of time before the CoM reaches the limit of the balance region.

A more suitable solution is moving the CoM to a position where the moment from gravity compensates for $\force$.
However, that solution cannot be generated by the WBQP since it does not immediately decrease the error on the force task. For example, when the robot is pushing a box, as in Fig.~\ref{fig:humanRobotInteraction} to generate enough force on the box, the robot needs to move its CoM forward. It has to decrease the pressure on the box, which is in opposition to reaching the target $\force$ and increasing the forces from the feet to move the CoM forward. Thus, a task is added for the WBQP to favor this behavior.

\subsection{Formulation of a CoM Admittance Task}
\label{sec:control_strategy_CoM_Admittance}

The admittance control is a common strategy to regulate dynamic force values for position-controlled robots, such as contact force~\cite{bouyarmane2019tro} or ZMP~\cite{caron2019icra}. We formulate a CoM admittance task applicable to a given measured force as,
\begin{equation}
  \label{eq:comAdmittance1}
  \comVel = \mathbf{K} (\force_{\text{tar}} - \force_{\text{mes}})
\end{equation}
where $\force_{\text{tar}}$ and $\force_{\text{mes}}$ are the target and measured contact forces in the global frame, respectively.
The objective is to express the matrix $\mathbf{K}$ such that the generated CoM velocity moves the CoM to decrease the force error.

Starting from equation~\eqref{eq:sum_moment} and considering no CoM acceleration, the moment generated only by the supposed contact and the momentum from gravity, we get:
\begin{equation}
  \label{eq:simplyfied_sum_moments}
  \mathbf{r} \times \force + \comPos \times m \gravity = \mathbf{0}
\end{equation}
where $r = [x, y, z]^T$ the contact position, $\force = [f_x, f_y, f_z]^T$ the contact force, $\comPos = [c_x, c_y, c_z]^T$ the CoM position and $\gravity = [0, 0, -g]^T$ with $g=9.81$~m.s$^{-2}$. Then by projecting on the $x$ and $y$ axes:
\begin{eqnarray}
  y f_z - z f_y &=& m c_y g\\
  z f_x - x f_z &=& - m c_x g
\end{eqnarray}
and writing,
\begin{equation}
  \Delta\force = \force_{\text{tar}} - \force_{\text{mes}} = \begin{bmatrix}
  \Delta f_x \\
  \Delta f_y \\
  \Delta f_z
  \end{bmatrix} \text{ and }  \Delta \comPos = \begin{bmatrix}
  \Delta c_x \\
  \Delta c_y \\ 
  \Delta c_z
  \end{bmatrix}
\end{equation}
where $\Delta \comPos$ is the desired CoM displacement, we derive:
\begin{eqnarray}
  \Delta c_y &=& \frac{y}{mg} \Delta f_z - \frac{z}{mg} \Delta f_y \\
  \Delta c_x &=& \frac{x}{mg} \Delta f_z - \frac{z}{mg} \Delta f_x 
\end{eqnarray}
then $\mathbf{K}$ is given by 
\begin{equation}
  \mathbf{K} = k_{\text{ad}}  \begin{bmatrix}
    -\frac{z}{mg} & 0 & \frac{y}{mg} \\
    0 & -\frac{z}{mg} & \frac{x}{mg} \\
    0 & 0 & 0
  \end{bmatrix}
  \label{eq:comAdmittanceMatrix}
\end{equation}
where $k_{\text{ad}}$ is tuned from experimental data.

The target CoM velocity computed in equation~\eqref{eq:comAdmittance1} is then employed as reference velocity in a CoM task with no stiffness but damping alone (which could be considered as a CoM velocity task). 


\section{Experiments}
\label{sec:experiments}

We assess our implementations with multi-contact experiments using the HRP-4 humanoid robot. Two experiments are conducted: (i) regulation of the right-hand contact force (Section~\ref{sec:experiment_regulation}) and (ii) pushing a $12.5$~kg box on a table (Section~\ref{sec:experiment_pushing}). A video of the following experiments is available online\footnotemark\footnotetext{\url{https://youtu.be/QLtO3rb4\_NY}}.

\subsection{Regulation of contact force}
\label{sec:experiment_regulation}

\begin{figure}[!ht]
  \centering
  \includegraphics[width=\columnwidth]{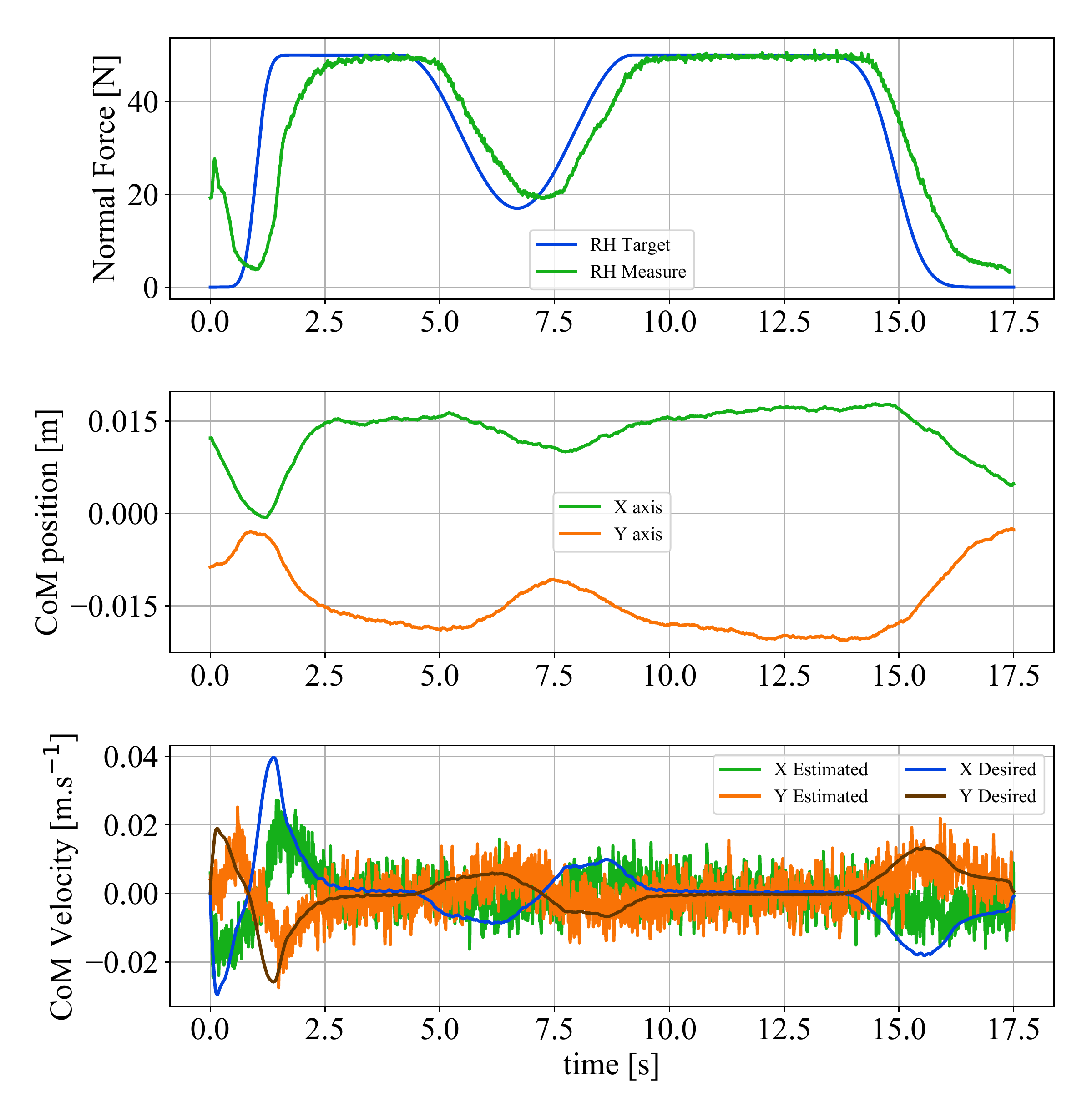}
  \caption{Right-hand force regulation. HRP-4 right hand is on a $70$~cm high horizontal table around $40$~cm in front of the standing robot and off centered by $30$~cm to the right. The left hand is not in contact.}
  \label{fig:rh_forcetracking}
\end{figure}

In this experiment, HRP-4 establishes contact on a table with its right hand, and the normal target force is smoothly increased to $50$~N. Then the desired force is set to a cosine curve down to $17$~N and back up to $50$~N. Finally, the normal target force is evenly decreased to $0$~N to remove the contact.
Figure~\ref{fig:rh_forcetracking} illustrates the tracking of the normal force as well as the CoM trajectory and velocity. 
Here the right-hand surface is horizontal, and the contact's normal is aligned with the gravity vector. The contact is in front of the robot and off-centered to the right. Thus, notice that when the target force increases and becomes higher than the measured force, the CoM moves forward and to the right. On the other hand, when the target force decreases and becomes lower than the measured force, the CoM moves backwards and to the left.  

\subsection{Box pushing in multi-contact}
\label{sec:experiment_pushing}

\begin{figure}[!ht]
  \centering
  \includegraphics[width=\columnwidth]{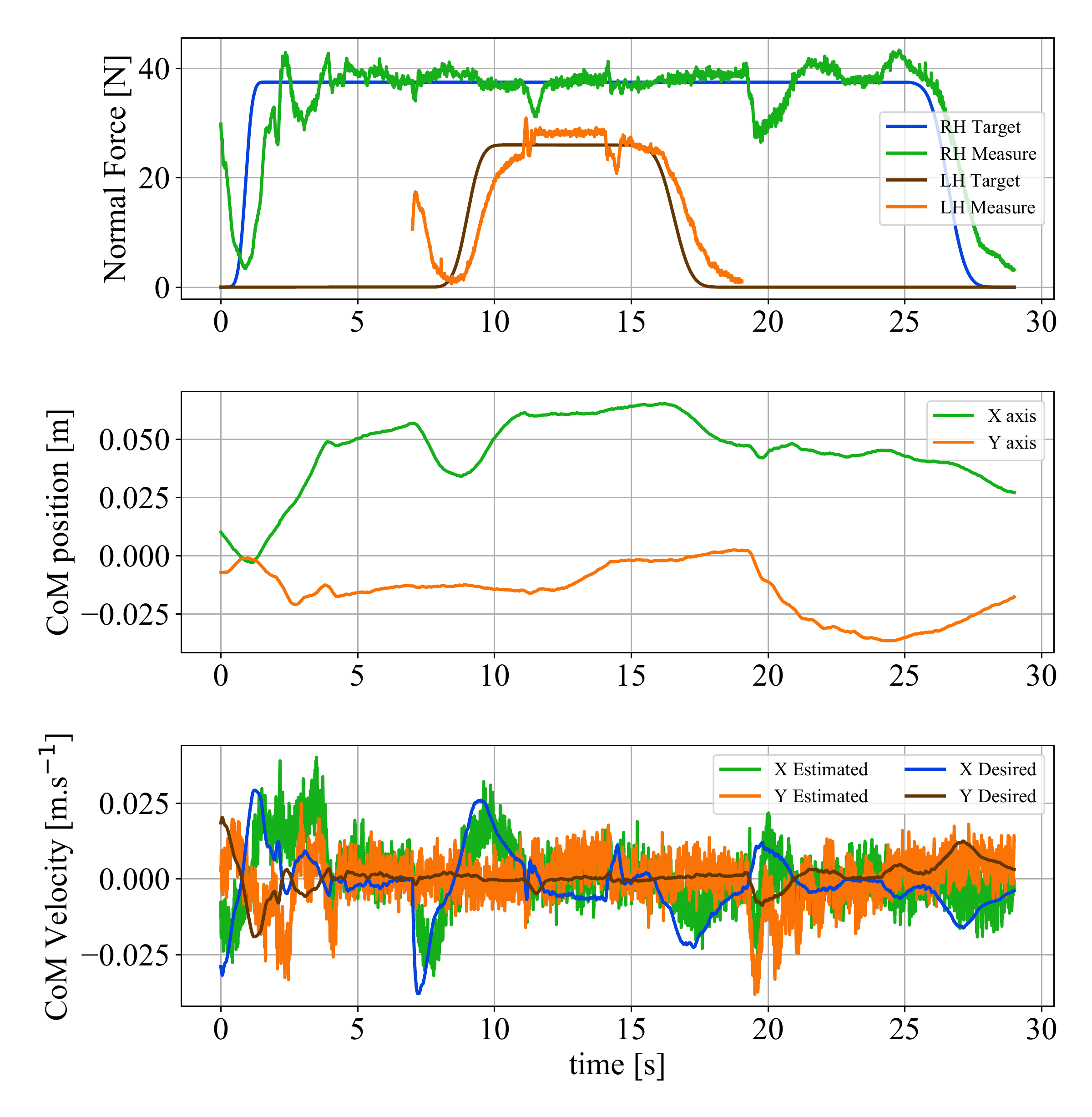}
  \caption{Box pushing in multi-contact (Fig~\ref{fig:humanRobotInteraction}): the right hand is in the same position as the first experiment (Fig~\ref{fig:rh_forcetracking}); the left hand is in contact with a box on the table. When the contact is established, the hand is $85$~cm high, $40$~cm in front of the robot and $30$~cm to its left. The box is $12.5$~kg, and during motion, it moves $15$~cm forward.}
  \label{fig:force_com_pushing_12_5}
\end{figure}

\begin{figure}[!ht]
  \centering
  \includegraphics[width=\columnwidth]{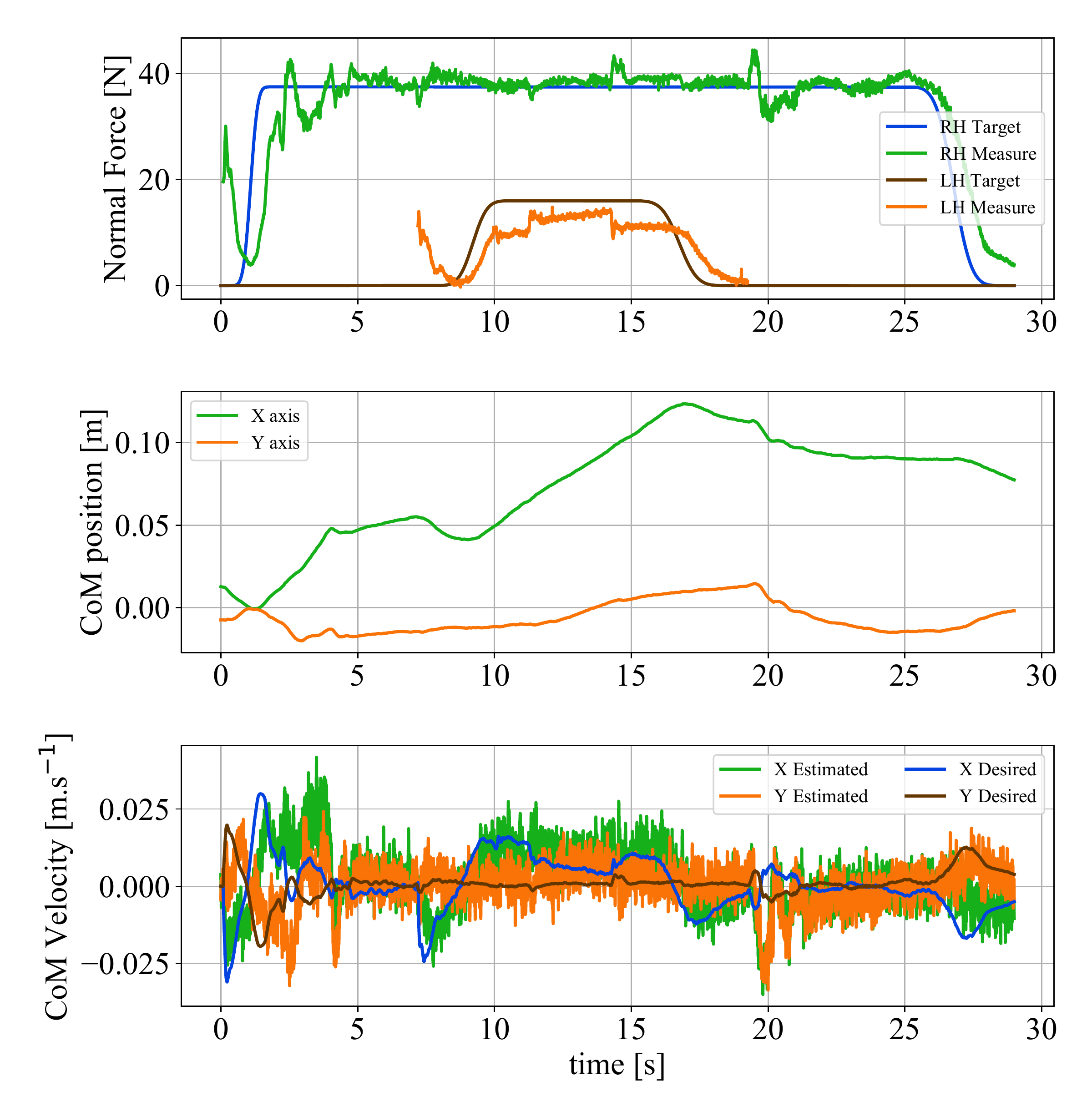}
  \caption{The same experiment as Fig~\ref{fig:force_com_pushing_12_5} with a box mass of $7.5$~kg: the target force on the left hand is not reached (since the box can be moved with a lower force); thus, the CoM keeps moving forward until the target is decreased to remove the contact.}
  \label{fig:force_com_pushing_7_5}
\end{figure}

Figure~\ref{fig:humanRobotInteraction} illustrates the second experiment. The robot first establishes contact with its right hand on the table as in the previous experiment, but regulate its force to $37.5$~N. Then it establishes contact with the box of mass $12.5$~kg using its left hand. The normal force is regulated to $26$~N. Then a position task is used to move the box $15$~cm forward. Ultimately, both contacts are removed (first left hand then right hand) after their normal forces have been decreased to $0$~N.
Figure~\ref{fig:force_com_pushing_12_5} shows the resulting force tracking, CoM trajectory and CoM velocity. The behavior concerning the right hand is similar to the previous scenario. However, for the left-hand contact, the contact normal is parallel to the global frame $x$-axis. Thus when controlling the normal force, motion of the CoM on the $y$-axis has no impact, and only motions along the $x$-axis are used (as a result of~\eqref{eq:comAdmittanceMatrix}). This is noticeable in the CoM velocity curve in Fig~\ref{fig:force_com_pushing_12_5}: from $7.5$ to $19$~s the desired velocity on the $y$-axis is almost zero while the velocity on the $x$-axis varies a lot. Notice that the regulation of the right hand is correct, and thus almost no CoM motion is generated by the right-hand contact.
Figure~\ref{fig:force_com_pushing_7_5} shows this experiment performed with a lighter box ($7.5$~kg).
In that case, the box starts to slide before the normal target force is reached. As a consequence, the CoM moved forward during the whole phase, where the left contact was established until the target was decreased. This happens because we do not identify the force required to move the box online. It is estimated from a prior experiment that in turn used an empirical value as target.

\subsection{Evolution of the balance region}

\begin{figure*}[!ht]
  \centering
  \includegraphics[height=2.8cm]{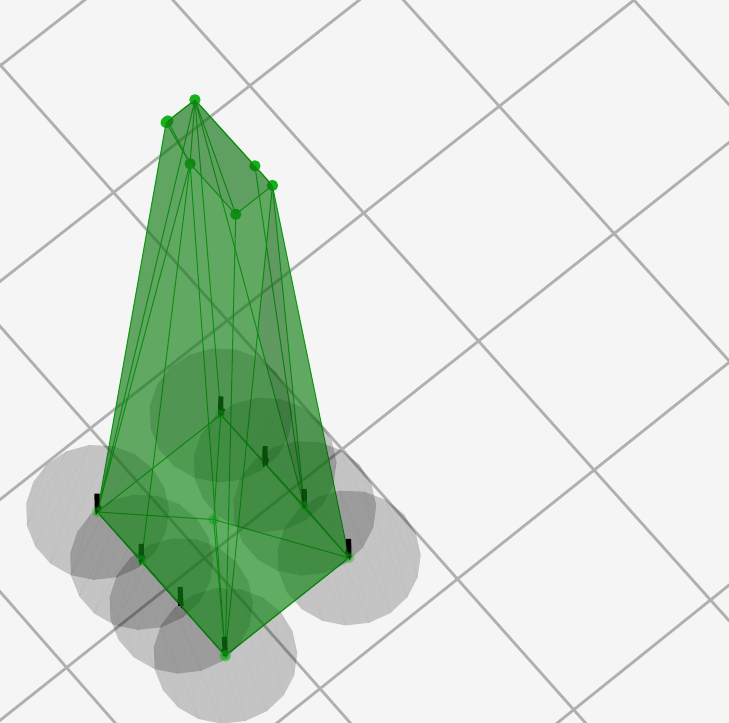}
  \includegraphics[height=2.8cm]{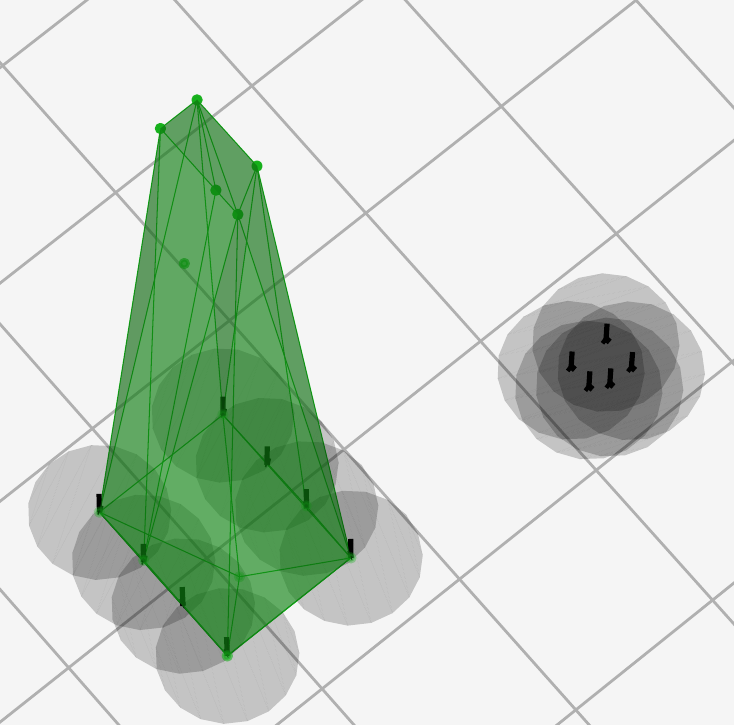}
  \includegraphics[height=2.8cm]{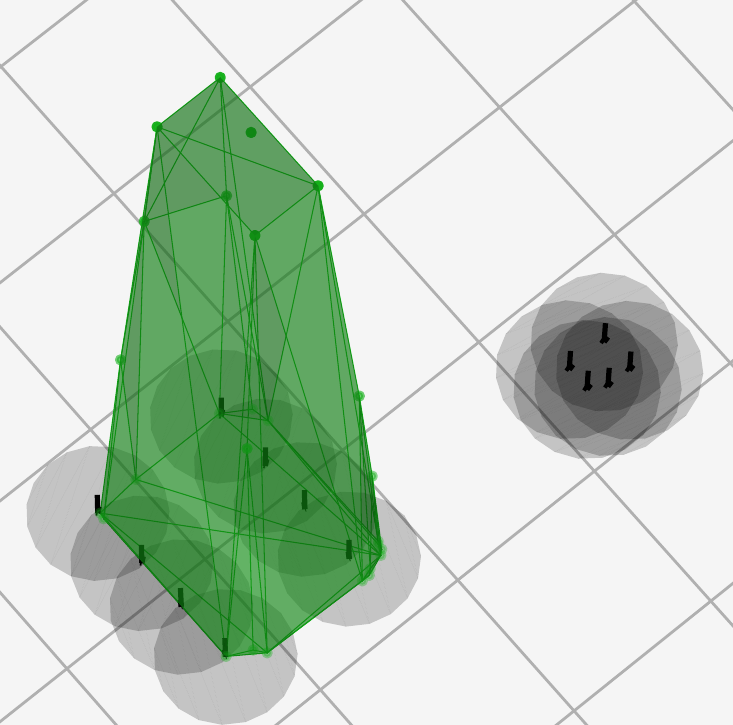}
  \includegraphics[height=2.8cm]{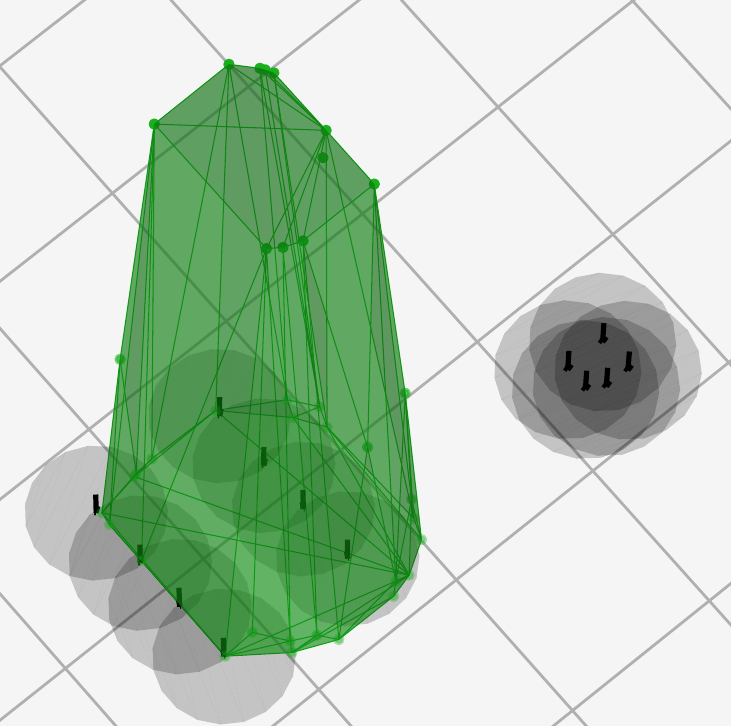}
  \includegraphics[height=2.8cm]{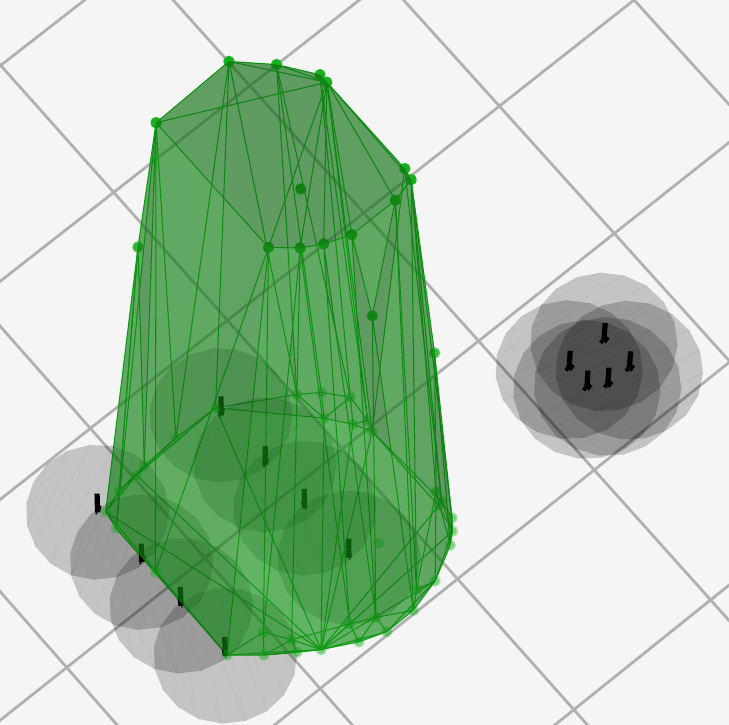}
  \includegraphics[height=2.8cm]{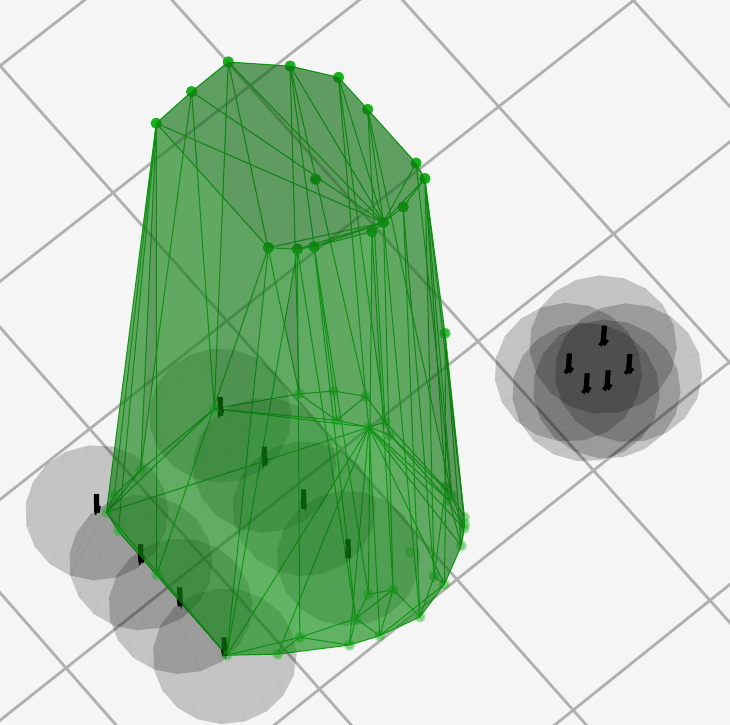}
  \caption{Evolution of the balance region when adding the right-hand contact with the table and increasing the normal force target. For the computation of the balance region, the contact normal force for each contact is bounded. Here, only the bounds on the right-hand contact are changing. From left to right: $[0.0,~0.0]$~[N](no right-hand contact), $[5e-4,~1e-2]$~[N], $[7.0,~20.5]$~[N], $[15.7,~47.2]$~[N], $[22.0,~66.5]$~[N], $[25.0,~75.0]$~[N](transition finished)}
  \label{fig:balance_region_evolution}
\end{figure*}

When there is a discrete change in the contact set, i.e. when a contact is added or removed, the balance region changes abruptly. To generate a smooth transition, the upper and lower bounds for each contact's normal force is increased or decreased progressively as suggested in~\cite{park2006icra}. This allows the balance region to evolve almost continuously. Figure~\ref{fig:balance_region_evolution} illustrate the evolution of the balance region when adding the right-hand contact.

Our task-based WBQP runs at $5$~ms. Meanwhile, the computation of the balance region takes between $10$~ms to $200$~ms (depending on the number of contacts). To avoid slowing down the calculation in the control loop, the computation of the balance region is done in a separate thread. Every time a balance region's computation has finished, the next one starts. As soon as a new region has been constructed, it is used to update the CoM constraint within the WBQP.

\section{Conclusion}
\label{sec:discussion_conclusion}

In this paper, we propose several ameliorations to the balance region algorithm proposed in~\cite{audren2018tro}.
These ameliorations enhanced the computational performances, allowing integrating balance regions directly as part of the task-space whole-body control framework. We also propose a CoM admittance strategy to improve the CoM behavior w.r.t. contact forces.
We assess our new method using multi-contacts, some fixed and others moving, experiments with the HRP-4 humanoid robot. 

As future work, we are focusing mainly on two streams. First, it will be challenging to improve further iterative methods to compute balance regions; we will instead investigate approaches from computational geometry that keep track of possible changes in the polytopes properties, e.g.~\cite{delos2015cmcgs,delos2015jamp}. Second, extensions to approximate shapes (Chebychev center)~\cite{samadi2021ral} could integrate the planning part of the CoM and the forces directly in the whole-body controller. This is possible if the balance region is formalized explicitly in the CoM acceleration space while considering possibly a set of Chebychev centers to counter conservativeness in elongated balance shapes.

\bibliographystyle{IEEEtran}
\bibliography{IEEEabrv,biblio}

\end{document}